\documentclass{llncs}

\usepackage{times}
\usepackage{graphicx} 
\usepackage{subfigure} 
\usepackage[numbers]{natbib}
\usepackage{algorithm}
\usepackage{algorithmic}
\usepackage{amsmath}
\usepackage{hyperref}
\usepackage{bm}
\usepackage{amssymb}

\newtheorem{lemma2}{Theorem}
\begin{document}

\title{Exact Hybrid Covariance Thresholding for Joint Graphical Lasso}

\author{Qingming Tang$^{\dagger}$ Chao Yang$^{\dagger}$ Jian Peng$^{\ddagger}$ Jinbo Xu$^{\dagger}$ }
\institute{ $^{\dagger}$Toyota Technological Institute at Chicago \\
\email{$\{$qmtang,harryyang,j3xu$\}$@ttic.edu} \\
$^{\ddagger}$ University of Illinois at Urbana-Champaign \\
\email{jianpeng@illinois.edu}}
\maketitle

\begin{abstract}

This paper studies precision matrix estimation for multiple related Gaussian graphical models from a dataset consisting of different classes, based upon the formulation of this problem as group graphical lasso. In particular, this paper proposes a novel hybrid covariance thresholding algorithm that can effectively identify zero entries in the precision matrices and split a large joint graphical lasso problem into many small subproblems. Our hybrid covariance thresholding method is superior to existing uniform thresholding methods in that our method can split the precision matrix of each individual class using different partition schemes and thus, split group graphical lasso into much smaller subproblems, each of which can be solved very fast. 
This paper also establishes necessary and sufficient conditions for our hybrid covariance thresholding algorithm. Experimental results on both synthetic and real data validate the superior performance of our thresholding method over the others. 

\end{abstract}

\section{Introduction}

Graphs have been widely used to describe the relationship between variables (or features). Estimating an undirected graphical model from a dataset has been extensively studied.  When the dataset has a Gaussian distribution, the problem is equivalent to estimating a precision matrix from the empirical (or sample) covariance matrix. In many real-world applications, the precision matrix is sparse. This problem can be formulated as graphical lasso \cite{banerjee2008model,yuan2007model} and many algorithms \cite{friedman2008sparse,rolfs2012iterative,hsieh2011sparse,witten2011new,tseng2009block} have been proposed to solve it. To take advantage of the sparsity of the precision matrix, some covariance thresholding (also called screening) methods are developed to detect zero entries in the matrix and then split the matrix into smaller submatrices, which can significantly speed up the process of estimating the entire precision matrix \cite{witten2011new,mazumder2012exact}.

Recently, there are a few studies on how to jointly estimate multiple related graphical models from a dataset with a few distinct class labels \cite{danaher2014joint,guo2011joint,hara2011common,honorio2010multi,liu2010efficient,mohan2012structured,mohan2014node,zhou2010time,zhu2014structural,yang2012fused,yuan2012alternating}. The underlying reason for joint estimation is that the graphs of these classes are similar to some degree, so it can increase statistical power and estimation accuracy by aggregating data of different classes. This joint graph estimation problem can be formulated as joint graphical lasso that makes use of similarity of the underlying graphs. In addition to group graphical lasso, Guo et al. used a non-convex hierarchical penalty to promote similar patterns among multiple graphical models~\cite{guo2011joint} ; \cite{danaher2014joint} introduced popular group and fused graphical lasso; and \cite{zhu2014structural,yang2012fused} proposed efficient algorithms to solve fused graphical lasso. To model gene networks, \cite{mohan2014node} proposed a node-based penalty to promote hub structure in a graph.

Existing algorithms for solving joint graphical lasso do not scale well with respect to the number of classes, denoted as $K$, and the number of variables, denoted as $p$. Similar to covariance thresholding methods for graphical lasso, a couple of thresholding methods \cite{zhu2014structural,yang2012fused} are developed to split a large joint graphical lasso problem into subproblems \cite{danaher2014joint}. Nevertheless, these algorithms all use uniform thresholding to decompose the precision matrices of distinct classes in exactly the same way. As such, it may not split the precision matrices into small enough submatrices especially when there are a large number of classes and/or the precision matrices have different sparsity patterns. Therefore, the speedup effect of covariance thresholding may not be very significant.

In contrast to the above-mentioned uniform covariance thresholding, this paper presents a novel hybrid (or non-uniform) thresholding approach that can divide the precision matrix for each individual class into smaller submatrices without requiring that the resultant partition schemes be exactly the same across all the classes. Using this method, we can split a large joint graphical lasso problem into much smaller subproblems. Then we employ the popular ADMM (Alternating Direction Method of Multipliers \cite{boyd2011distributed,gabay1976dual}) method to solve joint graphical lasso based upon this hybrid partition scheme. Experiments show that our method can solve group graphical lasso much more efficiently than uniform thresholding.

This hybrid thresholding approach is derived based upon group graphical lasso. The idea can also be generalized to other joint graphical lasso such as fused graphical lasso. Due to space limit, the proofs of some of the theorems in the paper are presented in supplementary material.

\section{Notation and Definition}

In this paper, we use a script letter, like $\mathcal{H}$, to denote a set or a set partition. When $\mathcal{H}$ is a set, we use $\mathcal{H}_i$ to denote the $i^{\text{th}}$ element. Similarly we use a bold letter, like $\bm{H}$ to denote a graph, a vector or a matrix. When $\bm{H}$ is a matrix we use $\bm{H}_{i,j}$ to denote its $(i,j)^{th}$ entry. We use $\{ \mathcal{H}^{(1)},\mathcal{H}^{(2)}, \ldots,\mathcal{H}^{(N)}\}$ and $\{\bm{H}^{(1)},\bm{H}^{(2)}\,\ldots,\bm{H}^{(N)}\}$ to denote $N$ objects of same category.

Let $\{\bm{X}^{(1)},\bm{X}^{(2)},\ldots,\bm{X}^{(K)} \}$ denote a sample dataset of $K$ classes and the data in $\bm{X}^{(k)}\ (1 \leq k \leq K)$ are independently and identically drawn from a $p$-dimension normal distribution $N(\bm{\mu} ^{(k)}, \bm{\Sigma}^{(k)})$. Let $\bm{S}^{(k)}$ and $\hat{\bm{\Theta}}^{(k)}$ denote the empirical covariance and (optimal) precision matrices of class $k$, respectively. By ``optimal'' we mean the precision matrices are obtained by exactly solving joint graphical lasso. Let a binary matrix $\bm{E}^{(k)}$ denote the sparsity pattern of $\hat{\bm{\Theta}}^{(k)}$, i.e., for any $i,j (1\leq i,j \leq p), \bm{E}_{i,j}^{(k)}=1$ if and only if $\hat{\bm{\Theta}}Θ_{i,j}^{(k)} \neq 0$.

\textbf{Set partition.} A set $\mathcal{H}$ is a partition of a set $\mathcal{C}$ when the following conditions are satisfied: 1) any element in $\mathcal{H}$ is a subset of $\mathcal{C}$; 2) the union of all the elements in $\mathcal{H}$ is equal to $\mathcal{C}$; and 3) any two elements in $\mathcal{H}$ are disjoint. Given two partitions $\mathcal{H}$ and $\mathcal{F}$ of a set $\mathcal{C}$, we say that $\mathcal{H}$ is \textbf{finer} than $\mathcal{F}$ (or $\mathcal{H}$ is a \textbf{refinement} of $\mathcal{F}$), denoted as $\mathcal{H} \preceq \mathcal{F}$, if every element in $\mathcal{H}$ is a subset of some element in $\mathcal{F}$. If $\mathcal{H} \preceq \mathcal{F}$ and $\mathcal{H} \neq \mathcal{F}$, we say that $\mathcal{H}$ is strictly finer than $\mathcal{F}$ (or $\mathcal{H}$ is a strict refinement of $\mathcal{F}$), denoted as $\mathcal{H} \prec \mathcal{F}$.

Let $\bm{\Theta}$ denote a matrix describing the pairwise relationship of elements in a set $\mathcal{C}$, where $\bm{\Theta}_{i,j}$ corresponds to two elements $\mathcal{C}_{i}$ and $\mathcal{C}_{j}$. Given a partition $\mathcal{H}$ of $\mathcal{C}$, we define $\bm{\Theta}_{\mathcal{H}_{k}}$as a $|\mathcal{H}_{k}| \times |\mathcal{H}_{k}|$ submatrix of $\bm{\Theta}$ where $\mathcal{H}_{k}$ is an element of $\mathcal{H}$ and
$(\bm{\Theta}_{\mathcal{H}_{k}})_{i,j} \cong   \bm{\Theta}_{(\mathcal{H}_{k})_{i}(\mathcal{H}_{k})_{j}}$ for any suitable ($i,j$).

\textbf{Graph-based partition.}	 Let $\mathcal{V}=\{1,2,\ldots,p\}$ denote the variable (or feature) set of the dataset. Let graph $ \bm{G}^{(k)}=(\mathcal{V} ,\bm{E}^{(k)})$ denote the $k^{\text{th}}$ estimated concentration graph $1 \leq k \leq K$. This graph defines a partition $\boxplus^{(k)}$  of $\mathcal{V}$, where an element in $\boxplus^{(k)}$ corresponds to a connected component in $\bm{G}^{(k)}$. The matrix $\hat{\bm{\Theta}}^{(k)}$ can be divided into disjoint submatrices based upon $\boxplus^{(k)}$. Let $\bm{E}$ denote the mix of $\bm{E}^{(1) },\bm{E}^{(2)}, \ldots, \bm{E}^{(K)}$, i.e., one entry $\bm{E}_{i,j}$ is equal to 1 if there exists at least one $k\ (1\leq k \leq K )$ such that $\bm{E}_{i,j}^{(k)}$ is equal to 1. We can construct a partition $\boxplus$ of $\mathcal{V}$ from graph $\bm{G}=\{\mathcal{V},\bm{E}\}$, where an element in $\boxplus$ corresponds to a connected component in $\bm{G}$. Obviously, $\boxplus^{(k)} \preccurlyeq \boxplus$ holds since $\bm{E}^{(k)}$ is a subset of $\bm{E}$. This implies that for any $k$, the matrix $\hat{\bm{\Theta}}^{(k)}$ can be divided into disjoint submatrices based upon $\boxplus$.

\textbf{Feasible partition.} A partition $\mathcal{H}$ of $\mathcal{V}$ is feasible for class $k$ or graph $\bm{G}^{(k)}$ if $\boxplus^{(k)} \preccurlyeq \mathcal{H}$. This implies that 1) $\mathcal{H}$ can be obtained by merging some elements in $\boxplus^{(k)}$; 2) each element in $\mathcal{H}$ corresponds to a union of some connected components in graph $\bm{G}^{(k)}$; and 3) we can divide the precision matrix $\hat{\bm{\Theta}}^{(k)}$ into independent submatrices according to $\mathcal{H}$ and then separately estimate the submatrices without losing accuracy. $\mathcal{H}$ is uniformly feasible if for all $k\ (1 \leq k \leq K)$, $\boxplus^{(k)} \preccurlyeq \mathcal{H}$ holds.

Let $\mathcal{H}^{(1)},\mathcal{H}^{(2)},\ldots,\mathcal{H}^{(K)}$ denote $K$ partitions of the variable set $V$. If for each $k\ (1 \leq k \leq K),\ \boxplus^{(k)} \preccurlyeq \mathcal{H}^{(k)}$ holds, we say $\{\mathcal{H}^{(1)}, \mathcal{H}^{(2) },\ldots, \mathcal{H}^{(K)} \}$ is a feasible partition of $\mathcal{V}$ for the $K$ classes or graphs. When at least two of the $K$ partitions are not same, we say $\{ \mathcal{H}^{(1)}, \mathcal{H}^{(2)},\ldots, \mathcal{H}^{(K)} \}$ is a non-uniform partition. Otherwise, $\{ \mathcal{H}^{(1)}, \mathcal{H}^{(2)},\ldots, \mathcal{H}^{(K)} \}$ is a class-independent or uniform partition and abbreviated as $\mathcal{H}$. That is, $\mathcal{H}$ is uniformly feasible if for all $k$ $(1\leq k \leq K)$, $\boxplus^{(k)} \preccurlyeq \mathcal{H}$ holds.
Obviously, $\{ \boxplus^{(1)}, \boxplus^{(2)}, \ldots,\boxplus^{(K)} \}$ is finer than any non-uniform feasible partition of the $K$ classes. Based upon the above definitions, we have the following theorem, which is proved in supplementary material.

\begin{lemma2}
 For any uniformly feasible partition $\mathcal{H}$ of the variable set $\mathcal{V}$, we have $\boxplus \preccurlyeq  \mathcal{H}$. That is, $\mathcal{H}$ is feasible for graph $\textbf{G}$ and $\boxplus$ is the finest uniform feasible partition.
\end{lemma2}

\begin{proof}
First, for any element $\mathcal{H}_j$ in $\mathcal{H}$, $\bm{G}$ does not contain edges between $\mathcal{H}_j$ and $\mathcal{H}-\mathcal{H}_j$. Otherwise, since $\bm{G}$ is the mixing (or union) of all $\bm{G}^{(k)}$, there exists at least one graph $\bm{G}^{(k)}$ such that it contains at least one edge between $\mathcal{H}_j$ and $\mathcal{H}-\mathcal{H}_j$. Since $\mathcal{H}_j$ is the union of some elements in $\boxplus^{(k)}$, this implies that there exist two different elements in $\boxplus^{(k)}$ such that $\bm{G}^{(k)}$ contains edges between them, which contradicts with the fact that $\bm{G}^{(k)}$ does not contain edges between any two elements in $\boxplus^{(k)}$. That is, $\mathcal{H}$ is feasible for graph $\bm{G}$. \par 
Second, if $\boxplus \preccurlyeq \mathcal{H}$ does not hold, then there is one element $\boxplus_i$ in $\boxplus$ and one element $\mathcal{H}_j$ in $\mathcal{H}$ such that $\boxplus_i \cap \mathcal{H}_j \neq \emptyset$ and $\boxplus_i-\mathcal{H}_j \neq \emptyset$. Based on the above paragraph, $\forall x \in \boxplus_i\cap \mathcal{H}_j$ and $\forall y \in \boxplus_i-\mathcal{H}_j=\boxplus_i\cap (\mathcal{H}_i-\mathcal{H}_j)$, we have $\bm{E}_{x,y}=\bm{E}_{y,x}=0$. That is, $\boxplus_i$ can be split into at least two disjoint subsets such that $\bm{G}$ does not contain any edges between them. This contradicts with the fact that $\boxplus_i$ corresponds to a connected component in graph $\bm{G}$.
\end{proof}

\section{Joint Graphical Lasso}

To learn the underlying graph structure of multiple classes simultaneously, some penalty functions are used to promote similar structural patterns among different classes, including \cite{rolfs2012iterative,danaher2014joint,guo2011joint,hara2011common,mohan2012structured,mohan2014node,zhu2014structural,yang2012fused,yuan2006model}. A typical joint graphical lasso is formulated as the following optimization problem:
\begin{equation}
\min \sum_{k=1}^{K} L({\bm{\Theta}}^{(k)}) + P(\bm{\Theta}) 
\end{equation}
Where $\bm{\Theta}^{(k)} \succ 0$ is the precision matrix $(k=1, \dots ,K)$ and $\bm{\Theta}$ represents the set of $\bm{\Theta}^{(k)}$. The negative log-likelihood $L(\bm{\Theta}^{(k)})$ and the regularization $P(\bm{\Theta})$  are defined as follows.
\begin{equation}
L(\bm{\Theta}^{(k)}) = -\log \det (\bm{\Theta}^{(k)}) + \mathrm{tr}(\mathcal{S}^{(k)}{\bm{\Theta}}^{(k)})
\end{equation}
\begin{equation}
P(\bm{\Theta}) = \lambda_{1}\sum_{k=1}^{K}\| \bm{\Theta}^{(k)}\|_{1} + \lambda_{2}J(\bm{\Theta})
\end{equation}

Here $\lambda_{1}>0$ and $\lambda_{2}>0$ and $J(\bm{\Theta})$ is some penalty function used to encourage similarity (of the structural patterns) among the $K$ classes. In this paper, we focus on group graphical lasso. That is,
\begin{eqnarray}
J(\bm{\Theta}) = 2\sum_{1 \leq i<j\leq p} \sqrt{\sum_{k=1}^{K}(\bm{\Theta}_{i,j}^{(k)})^{2}}
\end{eqnarray}

\section{Uniform Thresholding}
Covariance thresholding methods, which identify zero entries in a precision matrix before directly solving the optimization problem like Eq.(1), are widely used to accelerate solving graphical lasso. In particular, a screening method divides the variable set into some disjoint groups such that when two variables (or features) are not in the same group, their corresponding entry in the precision matrix is guaranteed to be 0. Using this method, the precision matrix can be split into some submatrices, each corresponding to one distinct group. To achieve the best computational efficiency, we shall divide the variable set into as small groups as possible subject to the constraint that two related variables shall be in the same group. Meanwhile, \cite{danaher2014joint} described a screening method for group graphical lasso. This method uses a single thresholding criterion (i.e., uniform thresholding) for all the $K$ classes, i.e., employs a uniformly feasible partition of the variable set across all the $K$ classes. Existing methods such as those described in \cite{danaher2014joint,zhu2014structural,yang2012fused} for fused graphical lasso and that in \cite{oztoprak2012newton} for node-based learning all employ uniform thresholding.

\begin{figure}[h]
\vspace{.3in}
\centering
\subfigure[]{\includegraphics[width=0.36\textwidth]{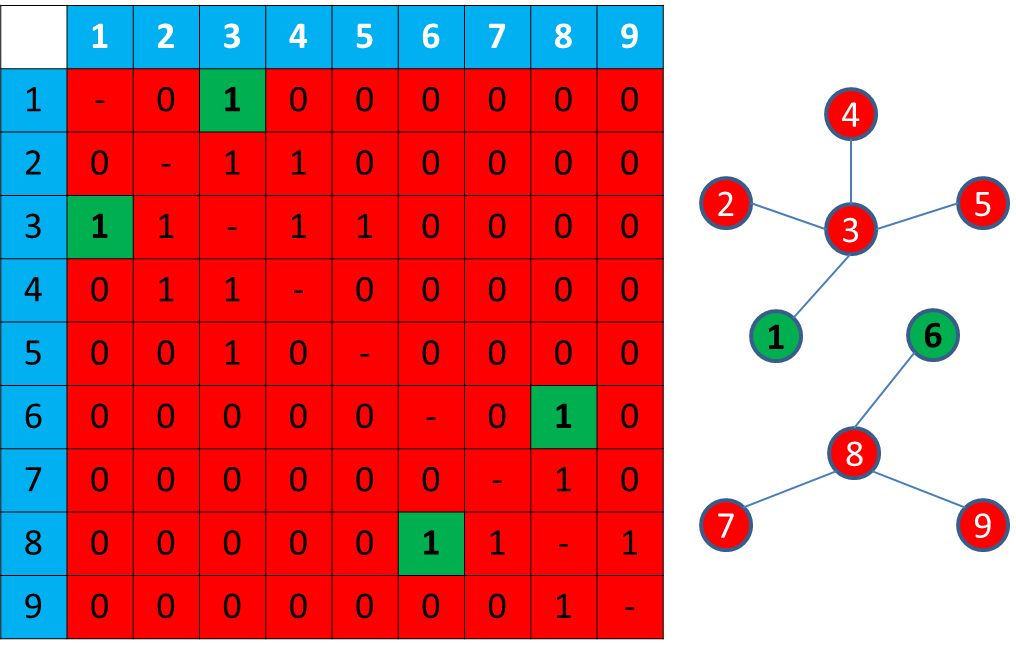}}
\subfigure[]{\includegraphics[width=0.18\textwidth]{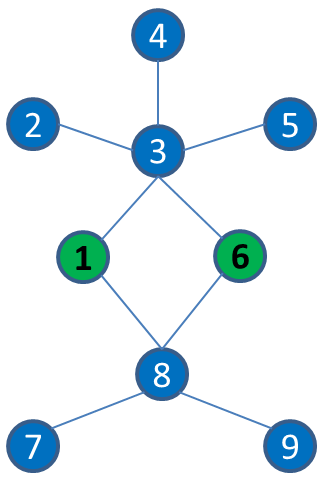}}
\subfigure[]{\includegraphics[width=0.36\textwidth]{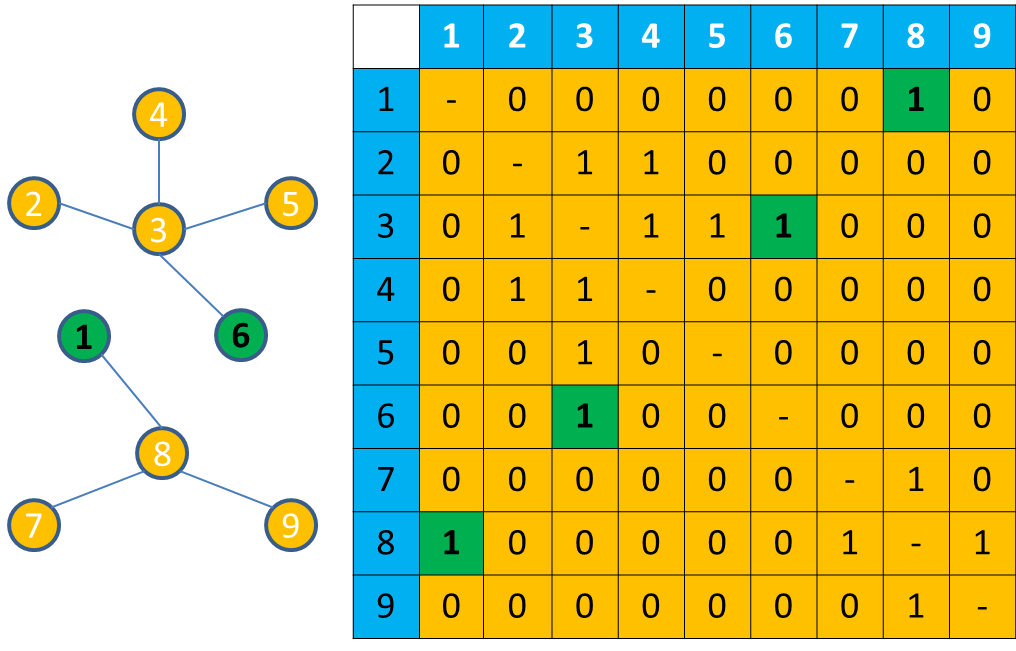}}
\caption{Illustration of uniform thresholding impacted by minor structure difference between two classes. (a) and (c): the edge matrix and concentration graph for each of the two classes. (b): the concentration graph resulting from the mixing of two graphs in (a) and (c).}
\label{drawbak_uniform}
\end{figure}

Uniform thresholding may not be able to divide the variable set into the finest feasible partition for each individual class when the $K$ underlying concentration graphs are not exactly the same. For example, Figure \ref{drawbak_uniform}(a) and (c) show two concentration graphs of two different classes. These two graphs differ in variables 1 and 6 and each graph can be split into two connected components. However, the mixing graph in (b) has only one connected component, so it cannot be split further. According to \textbf{Theorem 1}, no uniform feasible partition can divide the variable set into two disjoint groups without losing accuracy. It is expected that when the number of classes and variables increases, uniform thresholding may perform even worse.

\section{Non-uniform Thresholding}

\begin{figure}[t]
\centering
\includegraphics[width=0.9\columnwidth]{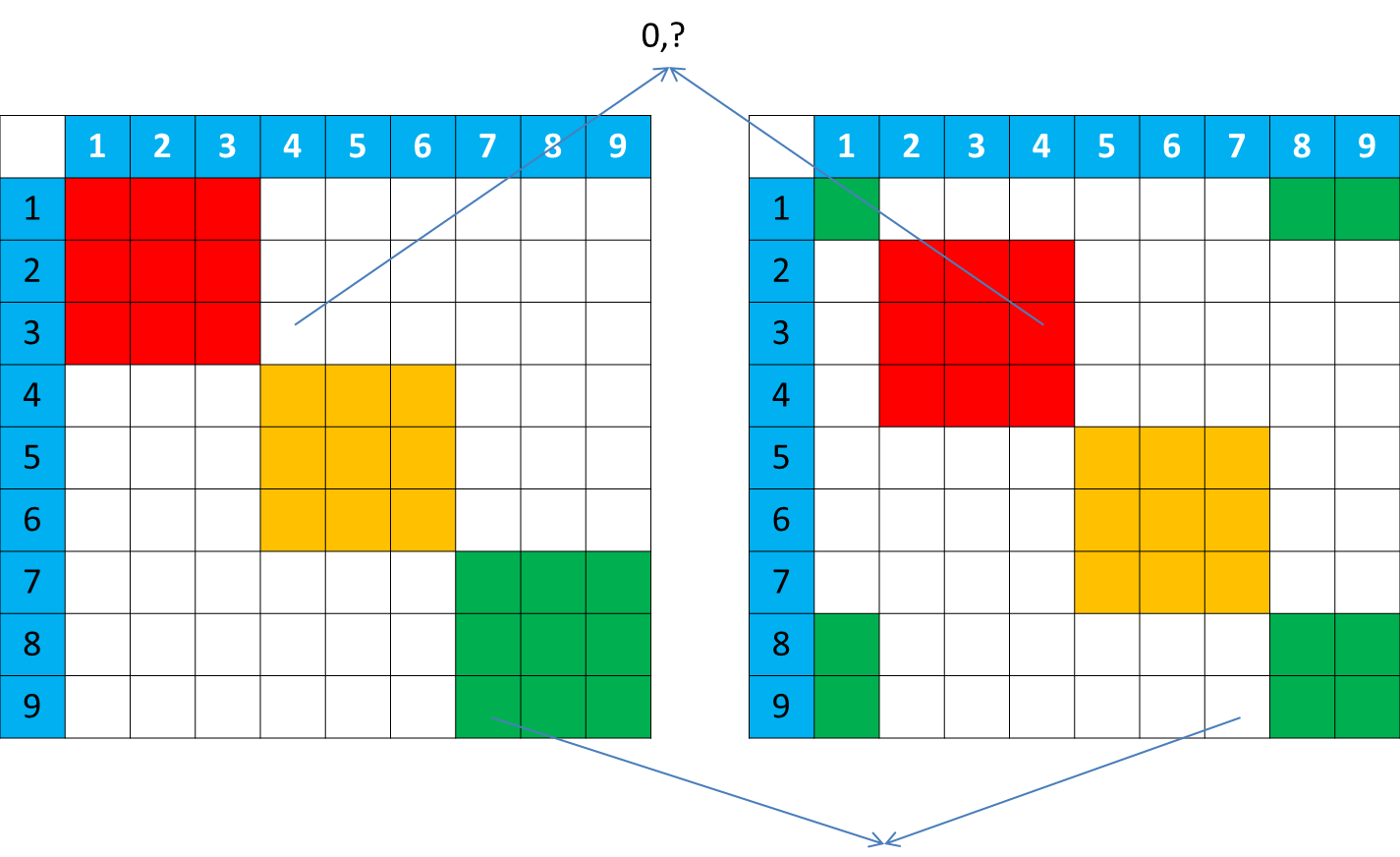}
\caption{Illustration of a non-uniform partition. White color indicates zero entries detected by covariance thresholding. Entries with the same color other than white belong to the same group.}
\label{illustrate_non_uniform}
\end{figure}

Non-uniform thresholding generates a non-uniform feasible partition by thresholding the $K$ empirical covariance matrices separately. In a non-uniform partition, two variables of the same group in one class may belong to different groups in another class. Figure \ref{illustrate_non_uniform} shows an example of non-uniform partition. In this example, all the matrix elements in white color are set to 0 by non-uniform thresholding. Except the white color, each of the other colors indicates one group. The $7^{\text{th}}$ and $9^{\text{th}}$ variables belong to the same group in the left matrix, but not in the right matrix. Similarly, the $3^{\text{rd}}$ and $4^{\text{th}}$ variables belong to the same group in the right matrix, but not in the left matrix.

We now present necessary and sufficient conditions for identifying a non-uniform feasible partition for group graphical lasso, with penalty defined in Eq (3) and (4).

Given a non-uniform partition $\{ \mathcal{P}^{(1)}, \mathcal{P}^{(2)}, \ldots, \mathcal{P}^{(K)} \}$ for the $K$ classes, let $F^{(k)}(i)=t$ denote the group which the variable $i$ belongs to in the $k^{\text{th}}$ class, i.e., $F^{(k)}(i) \Leftrightarrow i\in \text{P}_{t}^{(k)}$. We define pairwise relationship matrices $\textbf{I}^{(k)}\ (1 \leq k \leq K)$ as follows:
\begin{equation}
\begin{cases}
\textbf{I}_{i,j}^{(k)} = \textbf{I}_{j,i}^{(k)} = 0;\ \text{if} \ F^{(k)}(i) \neq F^{(k)}(j) \\
\textbf{I}_{i,j}^{(k)} = \textbf{I}_{j,i}^{(k)} = 1;\ \text{otherwise}
\end{cases}
\end{equation}

Also, we define $\bm{Z}^{(k)}(1 \leq k \leq K)$ as follows:
\begin{equation}
\bm{Z}_{i,j}^{(k)} = \bm{Z}_{j,i}^{(k)} = \lambda_{1} + \lambda_{2} \times \tau ((\sum_{t \neq k} |\hat{\bm{\Theta}}_{i,j}^{(t)}|) = 0)
\end{equation}
Here $\tau(b)$ is the indicator function.

The following two theorems state the necessary and sufficient conditions of a non-uniform feasible partition. See supplementary material for their proofs.

\begin{lemma2}
If $\{ \mathcal{P}^{(1)}, \mathcal{P}^{(2)}, \ldots, \mathcal{P}^{(K)} \} $ is a non-uniform feasible partition of the variable set $\mathcal{V}$, then for any pair $(i,j)\ (1 \leq i \neq j \leq p)$ the following conditions must be satisfied:
 \begin{equation}
 \begin{cases}
  \sum_{k=1}^{K}( |\mathcal{\textbf{S}}_{i,j}^{(k)}| - \lambda_{1})_{+}^{2}     \leq \lambda_{2}^{2}; \ \text{if}  \ \forall k \in 1,2, \ldots, K, \textbf{I}_{i,j}^{(k)} =0\\
  |\mathcal{\textbf{S}}_{i,j}^{(k)}| \leq \mathcal{\textbf{Z}}_{i,j}^{(k)}; \ \text{if} \ \textbf{I}_{i,j}^{(k)} =0 \ \text{and} \ \exists t \neq k, \textbf{I}_{i,j}^{(t)} = 1
 \end{cases}
 \end{equation}
Here, each $\mathcal{\textbf{S}}^{(k)}$ is a covariance matrix of the $k^{th}$ class and $x_{+} = \max(0, x)$.

\end{lemma2}

\begin{lemma2}
If for any pair $(i,j)(1 \leq i \neq j \leq p)$ the following conditions hold, then $\{\mathcal{P}^{(1)}, \mathcal{P}^{(2)}, \ldots, \mathcal{P}^{(K)} \}$ is a non-uniform feasible partition of the variable set $\mathcal{V}$.
 \begin{equation}
 \begin{cases}
  \sum_{k=1}^{K}( |\mathcal{\textbf{S}}_{i,j}^{(k)}| - \lambda_{1})_{+}^{2}     \leq \lambda_{2}^{2};\   \text{if}  \ \forall k \in 1,2, \ldots, K, \textbf{I}_{i,j}^{(k)} =0		\\
  |\mathcal{\textbf{S}}_{i,j}^{(k)}| \leq \lambda_{1}; \ \text{if} \ \textbf{I}_{i,j}^{(k)} =0 \ \text{and} \ \exists t \neq k, \textbf{I}_{i,j}^{(t)} = 1
 \end{cases}
 \end{equation}
\end{lemma2}
\begin{algorithm}[t]
\caption{Hybrid Covariance Screening Algorithm}
\begin{algorithmic}
\FOR{$k=1\ to\ K$}
 	\STATE Initialize $\textbf{I}_{i,j}^{(k)} = \textbf{I}_{j,i}^{(k)} = 1$, $\forall 1 \leq i < j \leq p$\;
 	\STATE Set $\textbf{I}_{i,j}^{(k)} = 0 $, if $|\bm{S}_{i,j}^{(k)}| \le \lambda_{1}$ and $i \neq j$\;
 	\STATE Set $\textbf{I}_{i,j}^{(k)} = 0 $, if $\sum_{k=1}^{K}(|\bm{S}_{i,j}^{(k)}| - \lambda_{1})_{+}^{2} \le \lambda_{2}^{2}$ and $i \neq j$\;
\ENDFOR

\FOR{$k=1\ to\ K$}
	\STATE Construct a graph $\bm{G}^{(k)}$ for $\mathcal{V}$ from $\bm{I}^{(k)}$\;
	\STATE Find connected components of $G^{(k)}$\;
\FOR{$\forall (i,j)\ in\ the\ same\ component\ of\ \bm{G}^{(k)}$}
 	\STATE Set $\bm{I}_{i,j}^{(k)}=\bm{I}_{j,i}^{(k)}=1$\;
 \ENDFOR
 \ENDFOR
\REPEAT
	\STATE Search for triple $(x,i,j)$ satisfying the following condition:\\
$\bm{I}_{i,j}^{(x)}=0$, $|\bm{S}_{i,j}^{(x)}|>\lambda_1$ and $\exists s$, s.t. $\bm{I}_{i,j}^{(s)}=1$ \;
	\IF{$\exists (x,i,j)$ satisfies the condition above}
		\STATE merge the two components of $\bm{G}^{(x)}$ that containing variable $i$ and $j$ into new component;\\
		\FOR{$\forall (m,n)$ in this new component}
			\STATE Set $\bm{I}_{m,n}^{(x)}=\bm{I}_{n,m}^{(x)}=1$;
		\ENDFOR
	\ENDIF
\UNTIL{No such kind of triple.}\\
\textbf{return} the connected components of each graph which define the non-uniform feasible solution\;
\end{algorithmic}
\end{algorithm}
\textbf{Algorithm 1} is a covariance thresholding algorithm that can identify a non-uniform feasible partition satisfying condition (8). We call \textbf{Algorithm 1} hybrid screening algorithm as it utilizes both class-specific thresholding (e.g. $|\mathcal{\textbf{S}}_{i,j}^{(k)}| \leq \lambda_{1}$ ) and global thresholding (e.g. $\sum_{k=1}^{K}( |\mathcal{\textbf{S}}_{i,j}^{(k)}| - \lambda_{1})_{+}^{2} \leq \lambda_{2}^{2}$ ) to identify a non-uniform partition. This hybrid screening algorithm can terminate rapidly on a typical Linux machine, tested on the synthetic data described in section 7 with $K=10$ and $p=10000$.

We can generate a uniform feasible partition using only the global thresholding and generate a non-uniform feasible partition by using only the class-specific thresholding, but such a partition is not as good as using the hybrid thresholding algorithm. Let $\{ \mathcal{H}^{(1) },\mathcal{H}^{(2) }, \ldots,\mathcal{H}^{(K) } \}$ ,  $\{\mathcal{L}^{(1)}, \mathcal{L}^{(2)},\ldots,\mathcal{L}^{(K)} \}$  and $\mathcal{G}$ denote the partitions generated by hybrid, class-specific and global thresholding algorithms, respectively. It is obvious that $\mathcal{H}^{(k)} \preccurlyeq \mathcal{L}^{(k)}$ and $\mathcal{H}^{(k)} \preccurlyeq \mathcal{G}$ for $k=1,2,\dots,K$ since condition (8) is a combination of both global thresholding and class-specific thresholding.

\begin{figure}[t]
\centering
\includegraphics[width=0.9\columnwidth]{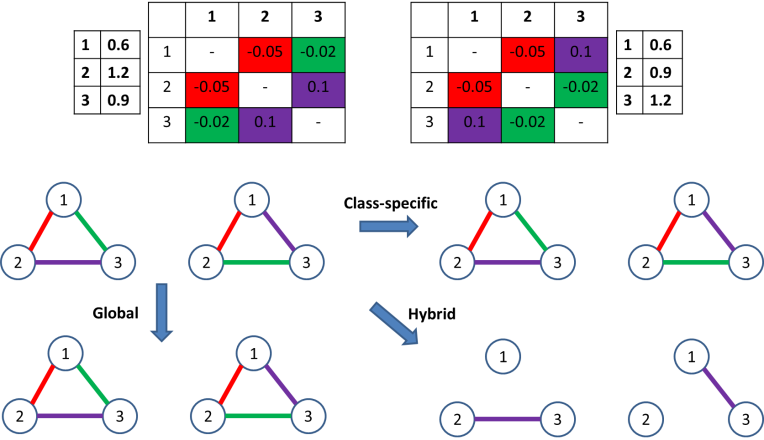}
\caption{Comparison of three thresholding strategies. The dataset contains 2 slightly different classes and 3 variables. The two sample covariance matrices are shown on the top of the figure. The parameters used are $\lambda_1=0.04$ and $\lambda_2=0.02$.}
\label{toy_example}
\end{figure}

Figure \ref{toy_example} shows a toy example comparing the three screening methods using a dataset of two classes and three variables. In this example, the class-specific or the global thresholding alone cannot divide the variable set into disjoint groups, but their combination can do so.

We have the following theorem regarding our hybrid thresholding algorithm, which will be proved in Supplemental File.

\begin{lemma2}
The hybrid screening algorithm yields the finest non-uniform feasible partition satisfying condition (8).
\end{lemma2}

\section{Hybrid ADMM (HADMM)}
In this section, we describe how to apply ADMM (Alternating Direction Method of Multipliers \cite{boyd2011distributed,gabay1976dual}) to solve joint graphical lasso based upon a non-uniform feasible partition of the variable set. According to \cite{danaher2014joint}, solving Eq.(1) by ADMM is equivalent to minimizing the following scaled augmented Lagrangian form:
\begin{equation}
\begin{aligned}
\sum\limits_{k=1}^KL(\bm{\Theta}^{(k)})+\frac{\rho}{2}\sum_{k=1}^{K}\| \bm{\Theta}^{(k)} - \bm{Y}^{(k)} + \bm{U}^{(k)}\|_{F}^{2}+P(\bm{Y})
\end{aligned}
\end{equation}
where $\bm{Y} = \{\bm{Y}^{(1)}, \bm{Y}^{(1)}, \ldots, \bm{Y}^{(K)} \}$ and $\bm{U} = \{ \bm{U}^{(1)}, \bm{U}^{(1)}, \ldots, \bm{U}^{(K)} \}$ are dual variables. We use the ADMM algorithm to solve Eq.(9) iteratively, which updates the three variables $\bm{\Theta}$, $\bm{Y}$ and $\bm{U}$ alternatively. The most computational-insensitive step is to update $\bm{\Theta}$ given $\bm{Y}$ and $\bm{U}$, which requires eigen-decomposition of $K$ matrices. We can do this based upon a non-uniform feasible partition $\{ \mathcal{H}^{(1)}, \mathcal{H}^{(2)}, \ldots, \mathcal{H}^{(K)} \}$. For each $k$, updating $\bm{\Theta}^{(k)}$ given $\bm{Y}^{(k)}$  and $\bm{U}^{(k)}$ for Eq (9) is equivalent to solving in total $|\mathcal{H}^{(k)}|$ independent sub-problems. For each $\mathcal{H}_j^{(k)}\in\mathcal{H}^{(k)}$, its independent sub-problem solves the following equation:
\begin{equation}
(\bm{\Theta}^{(k)}_{H_j^{(k)}})^{-1}=\mathcal{S}_{\mathcal{H}_{j}^{(k)}}^{(k)} + \rho \times (\bm{\Theta}_{\mathcal{H}_{j}^{(k)}}^{(k)} - \bm{Y}_{\mathcal{H}_{j}^{(k)}}^{(k)} + \bm{U}_{\mathcal{H}_{j}^{(k)}}^{(k)})
\end{equation}
Solving Eq.(10) requires eigen-decomposition of small submatrices, which shall be much faster than the eigen-decomposition of the original large matrices. Based upon our non-uniform partition, updating $\bm{Y}$ given $\bm{\Theta}$ and $\bm{U}$ and updating $\bm{U}$ given $\bm{Y}$ and $\bm{\Theta}$ are also faster than the corresponding components of the plain ADMM algorithm described in \cite{danaher2014joint}, since our non-uniform thresholding algorithm can detect many more zero entries before ADMM is applied.
\begin{figure}[t]
\centering
\includegraphics[width=0.85\columnwidth]{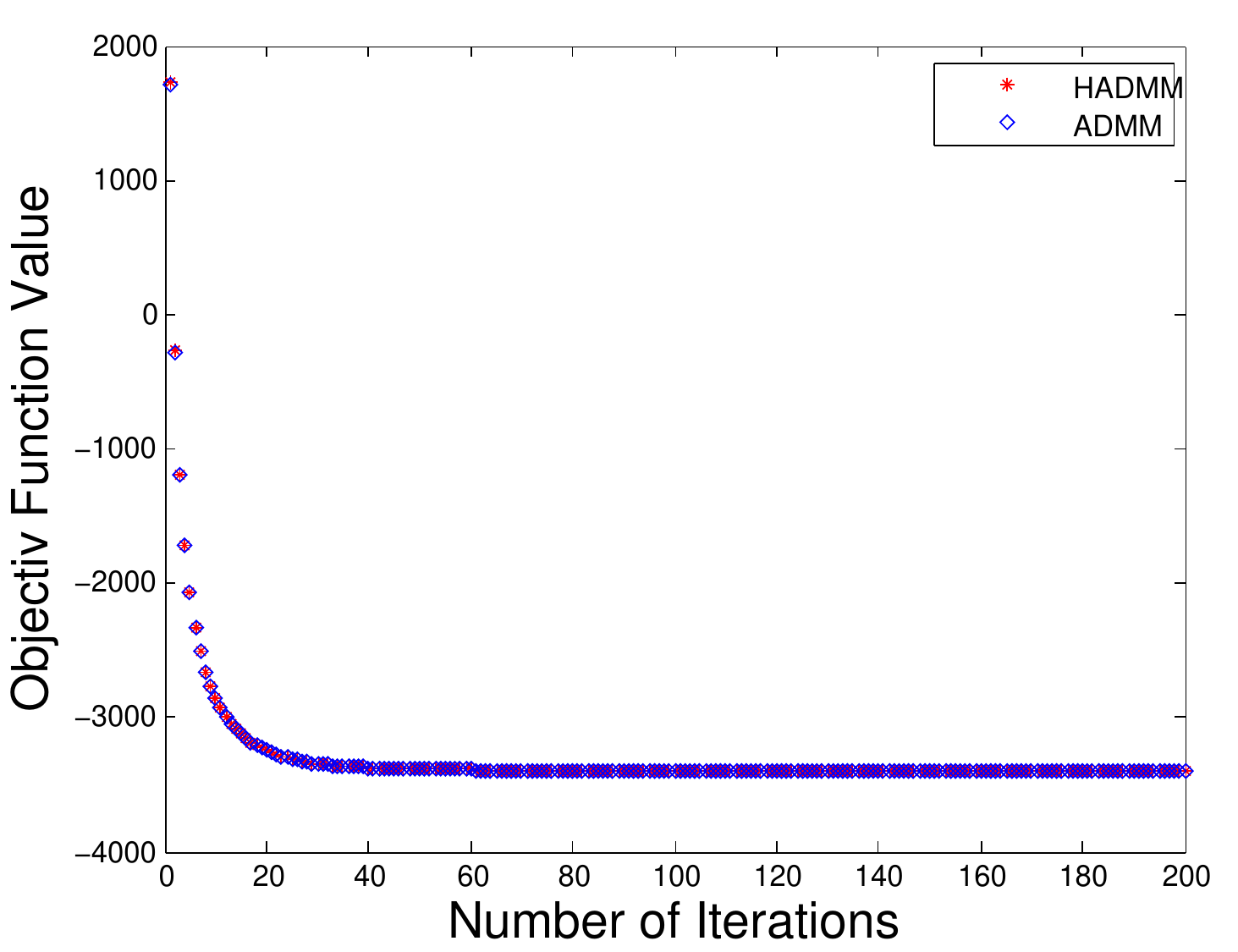}
\caption{The objective function value with respect to the number of iterations on a six classes type C data with $p=1000$, $\lambda_1=0.0082$ and $\lambda_2=0.0015$.}
\label{correctness}
\end{figure}

\section{Experimental Results}
We tested our method, denoted as HADMM (i.e., hybrid covariance thresholding algorithm + ADMM), on both synthetic and real data and compared HADMM with two control methods: 1) GADMM: global covariance thresholding algorithm + ADMM; and 2) LADMM: class-specific covariance thresholding algorithm +ADMM.
We implemented these methods with C++ and R, and tested them on a Linux machine with Intel Xeon E5-2670 2.6GHz. 

To generate a dataset with $K$ classes from Gaussian distribution, we first randomly generate $K$ precision matrices and then use them to sample $5 \times p$ data points for each class. To make sure that the randomly-generated precision matrices are positive definite, we set all the diagonal entries to 5.0, and an off-diagonal entry to either 0 or $ \pm r \times 5.0$ . We generate three types of datasets as follows.
\begin{itemize}
\item\textbf{Type A}: 97\% of the entries in a precision matrix are 0.
\item	\textbf{Type B}: the $K$ precision matrices have same diagonal block structure.	
\item	\textbf{Type C}: the $K$ precision matrices have slightly different diagonal block structures.
\end{itemize}
For \textbf{Type A}, $r$ is set to be less than 0.0061. For \textbf{Type B} and \textbf{Type C}, $r$ is smaller than 0.0067. For each type we generate 18 datasets by setting $K=2,3,\ldots,10$, and $p=1000,\ 10000$, respectively.

\subsection{Correctness of HADMM by Experimental Validation}
We first show that HADMM can converge to the same solution obtained by the plain ADMM (i.e., ADMM without any covariance thresholding) through experiments. 

\begin{figure*}[t]
 \centering
 \subfigure[\textbf{Type A}]{  \centering \includegraphics[width=0.31\textwidth]{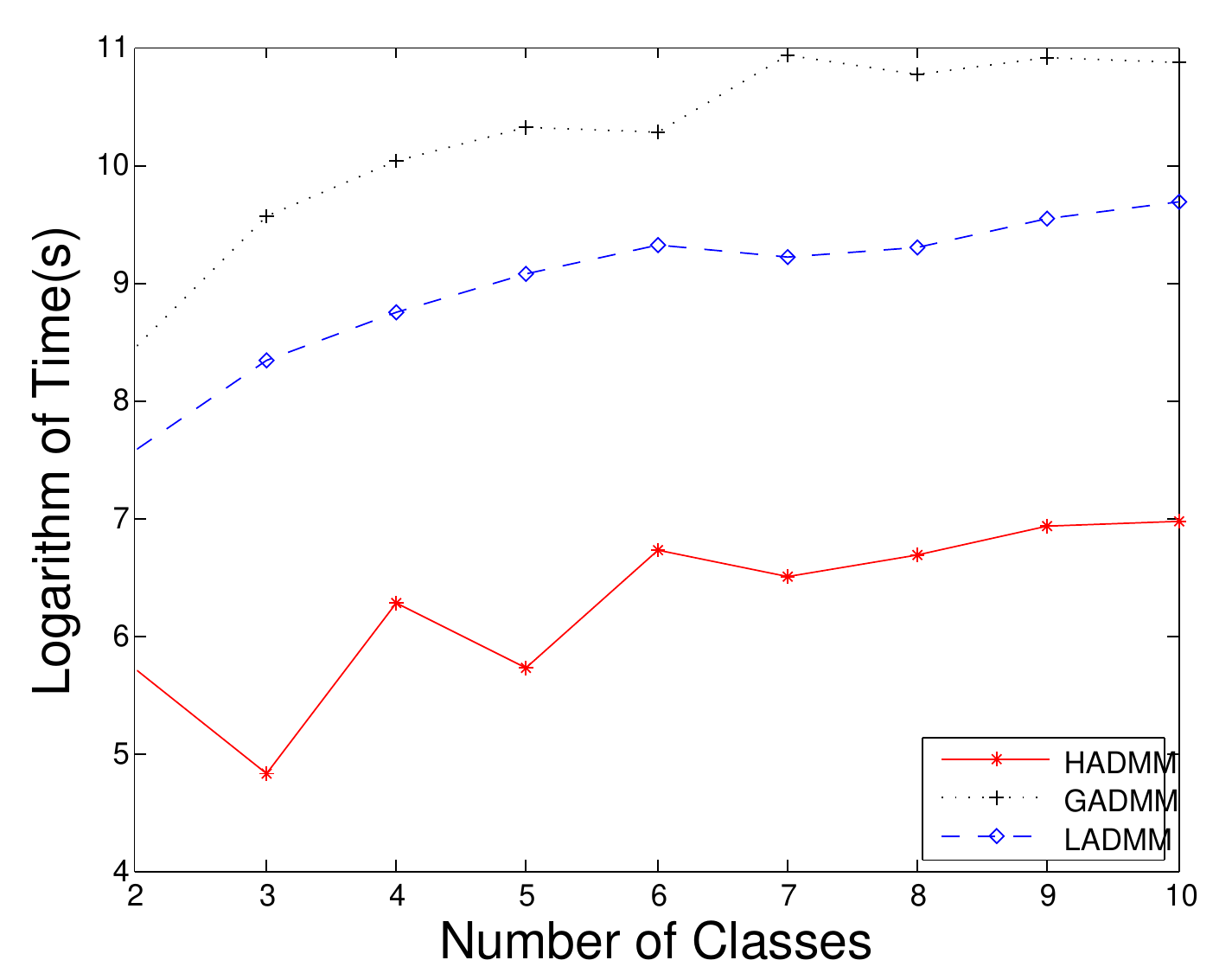}
 \includegraphics[width=0.31\textwidth]{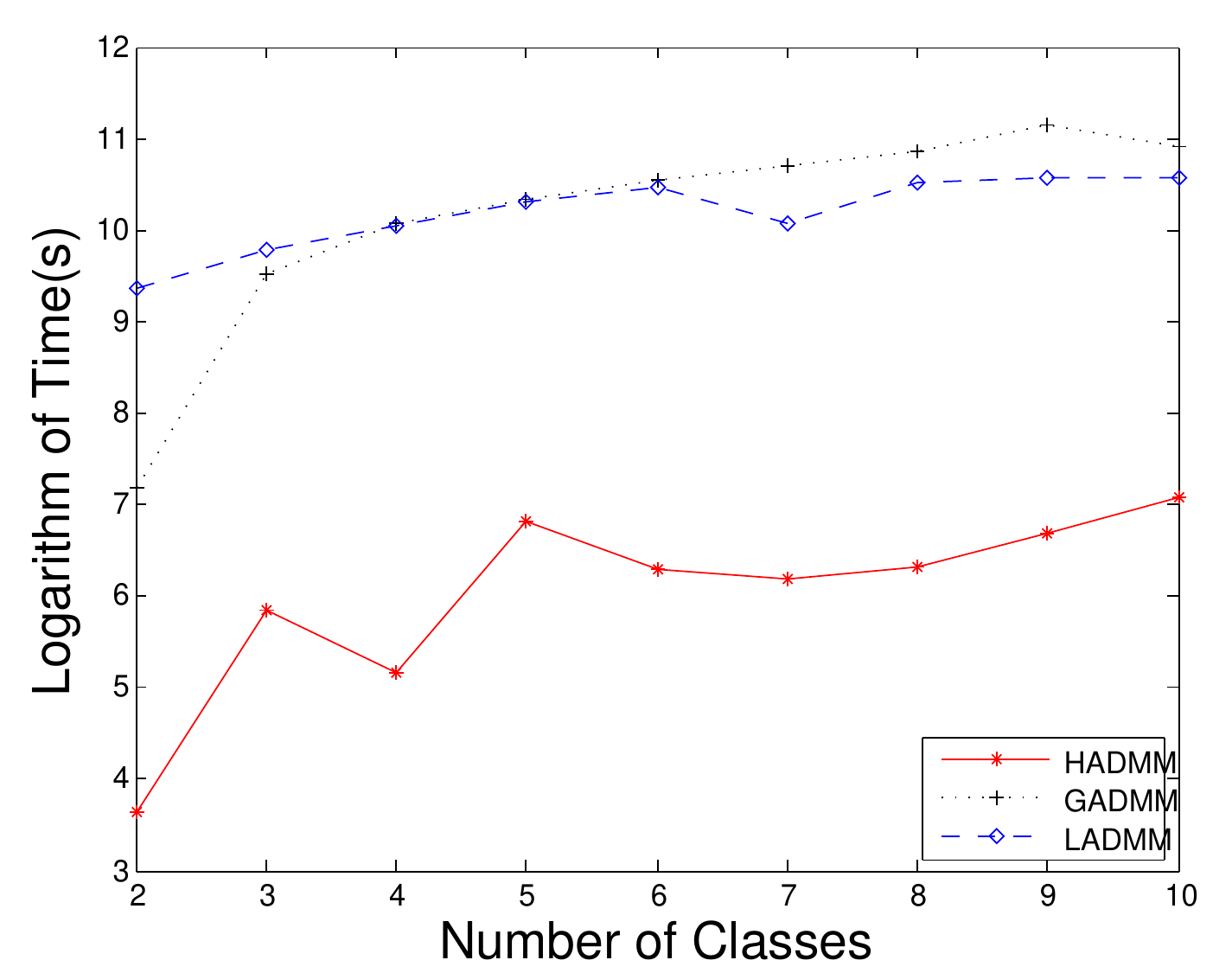}
 \includegraphics[width=0.31\textwidth]{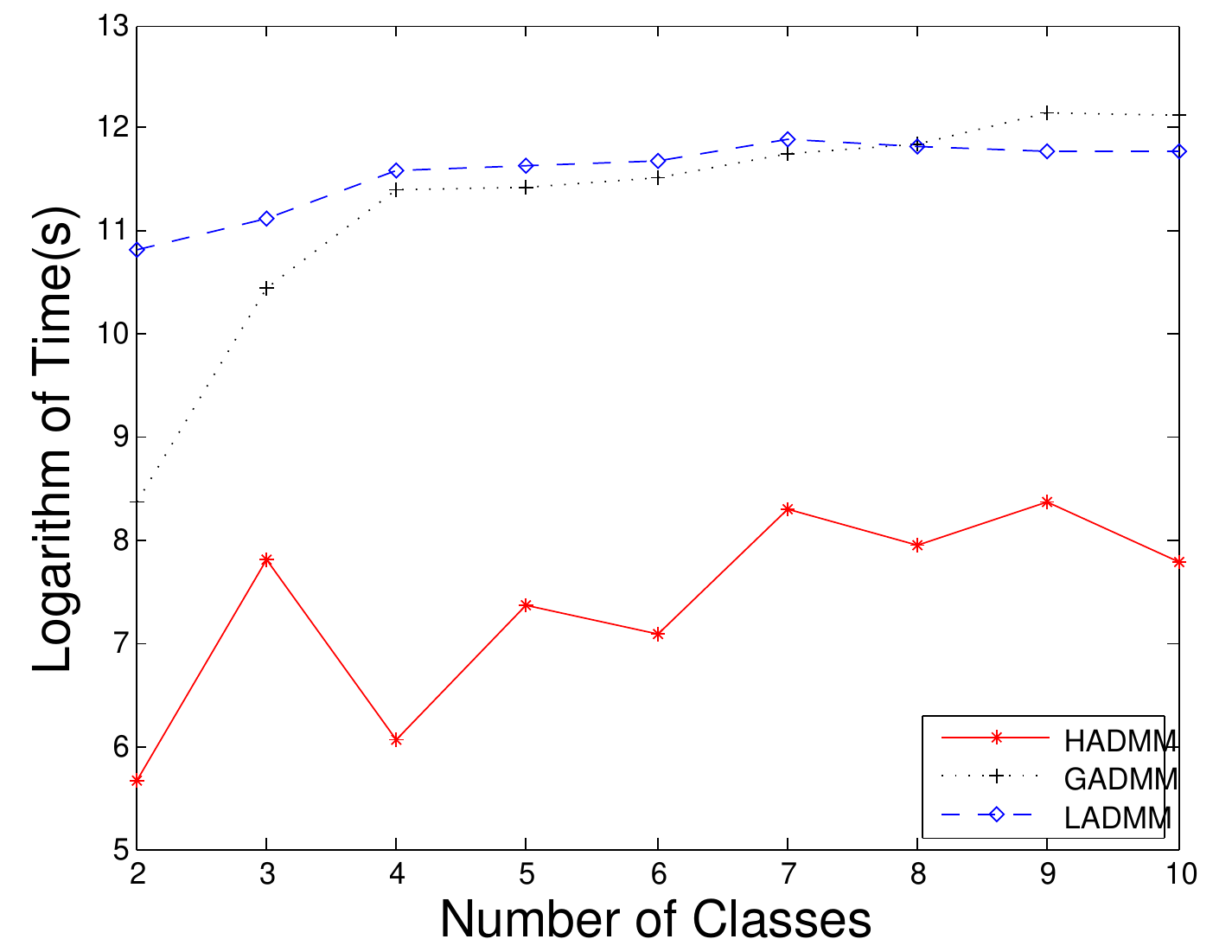}
 }
  \subfigure[\textbf{Type B}]{   \centering \includegraphics[width=0.31\textwidth]{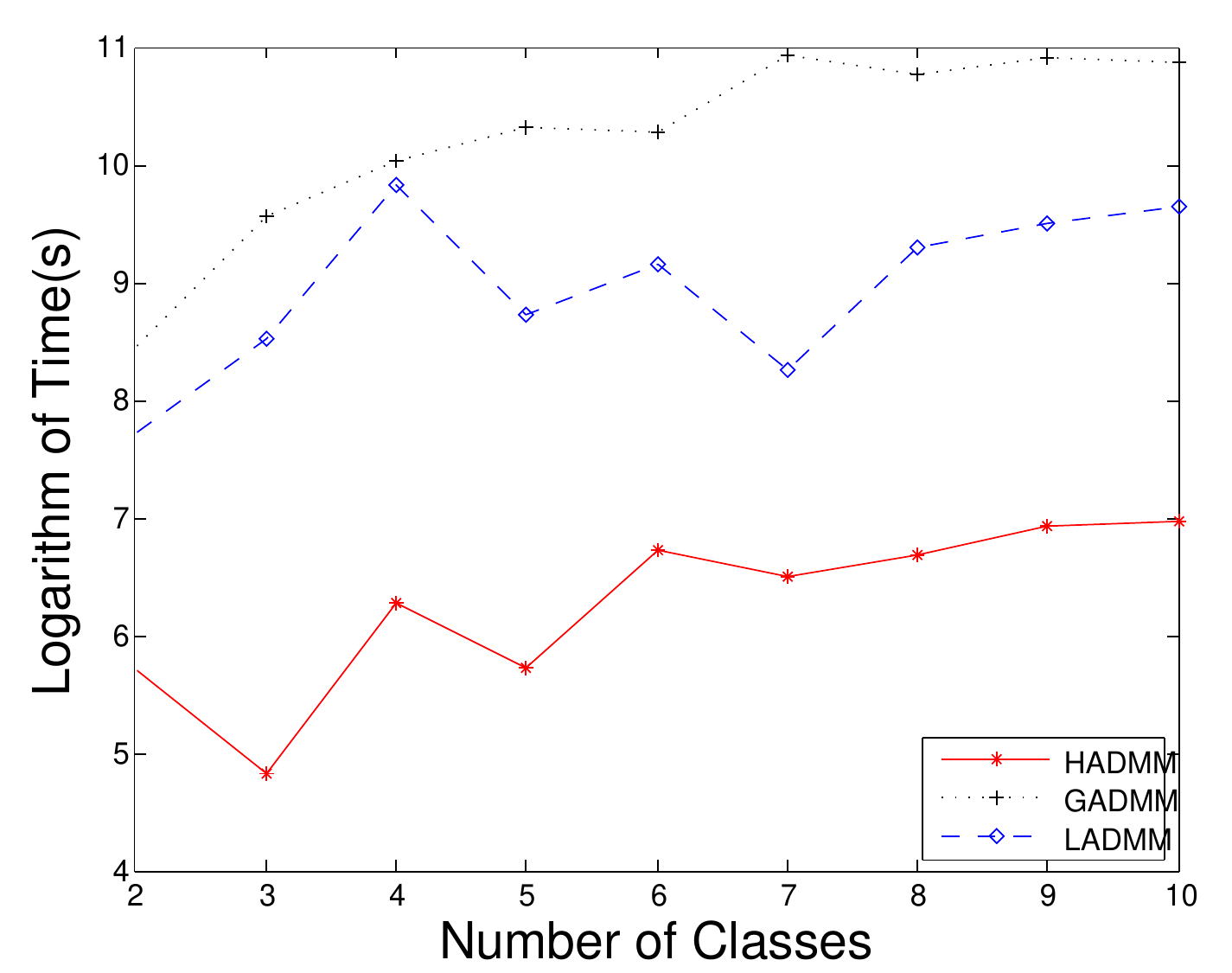}
 \includegraphics[width=0.31\textwidth]{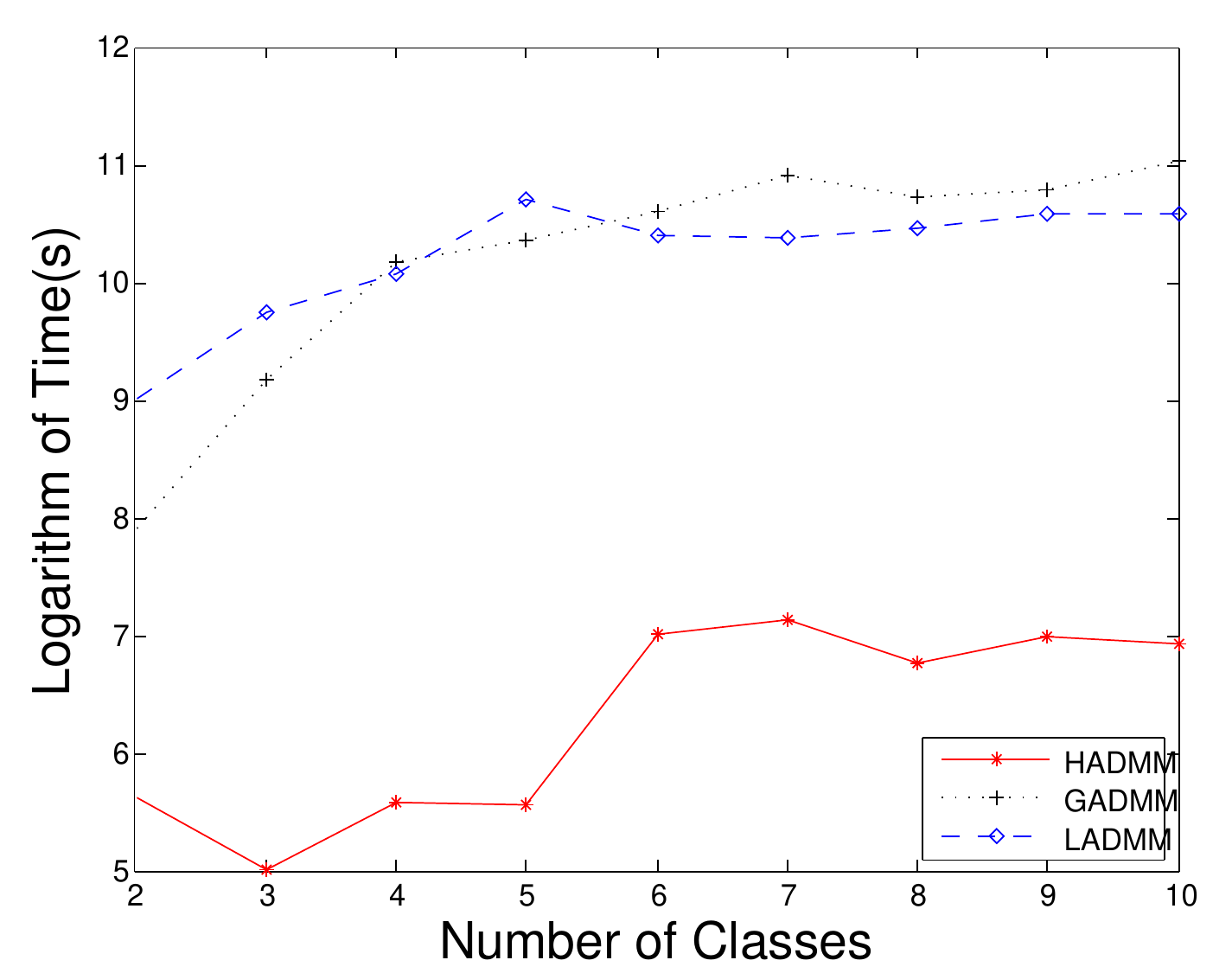}
 \includegraphics[width=0.31\textwidth]{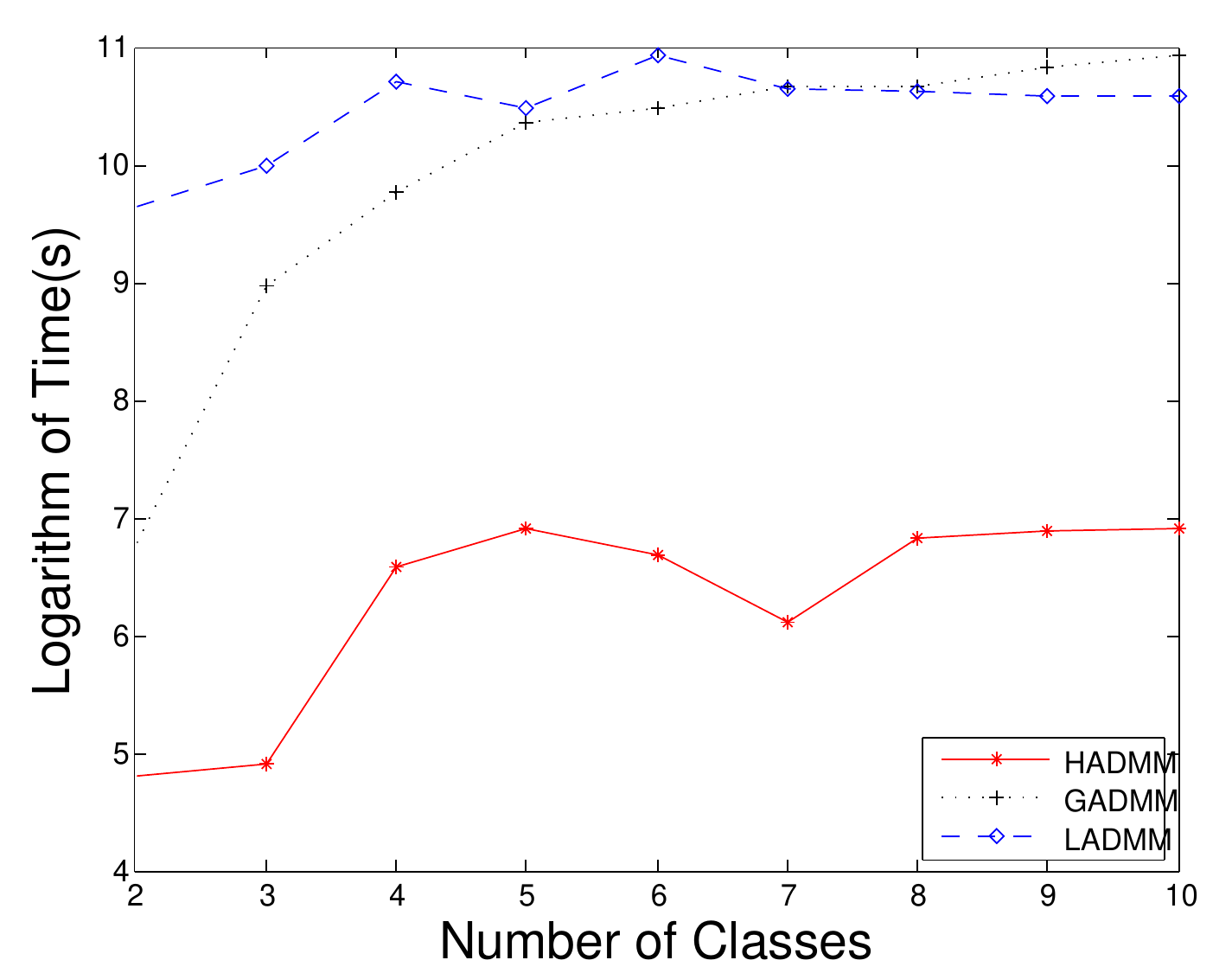}
 }
  \subfigure[\textbf{Type C}]{   \centering
 \includegraphics[width=0.31\textwidth]{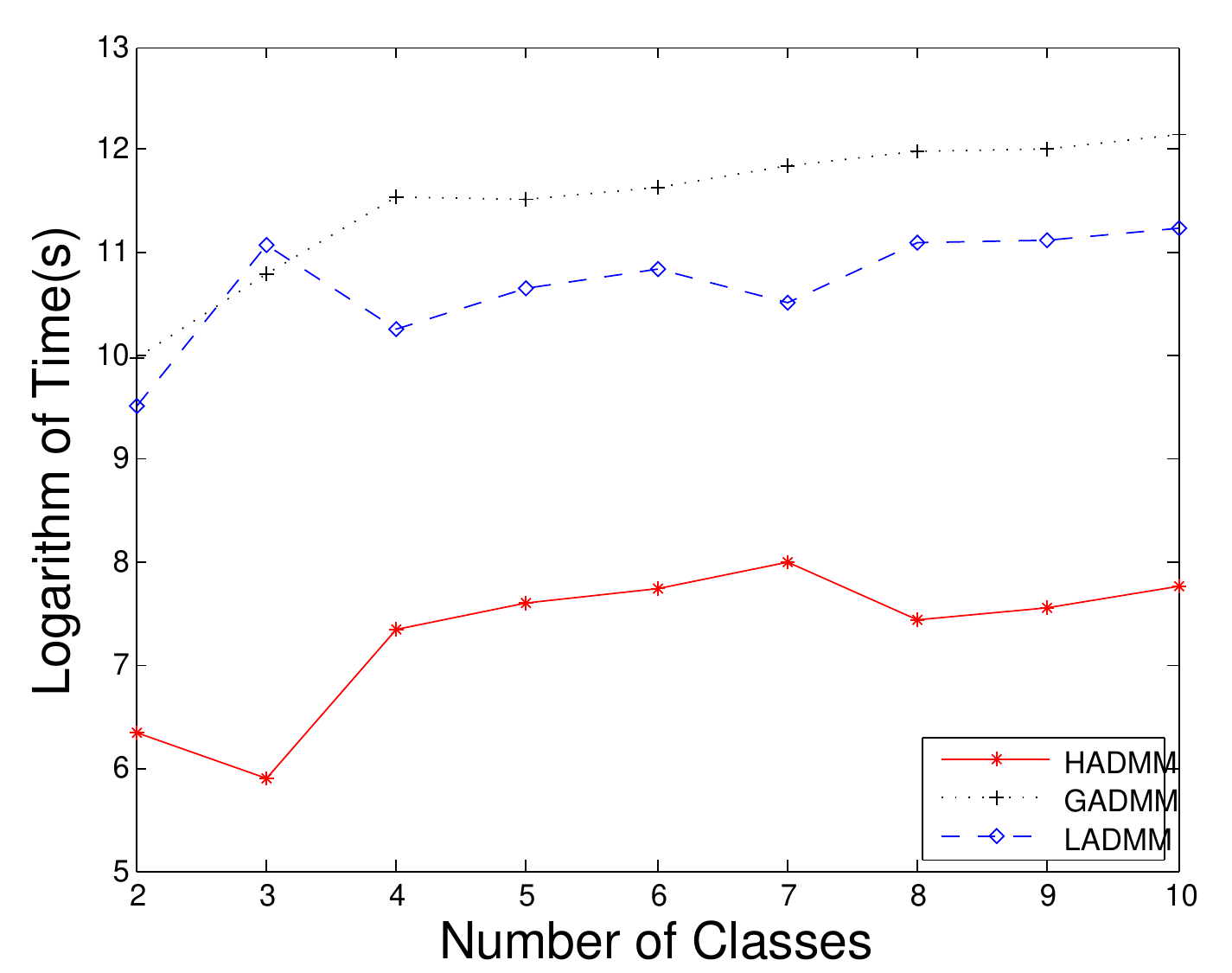}
 \includegraphics[width=0.31\textwidth]{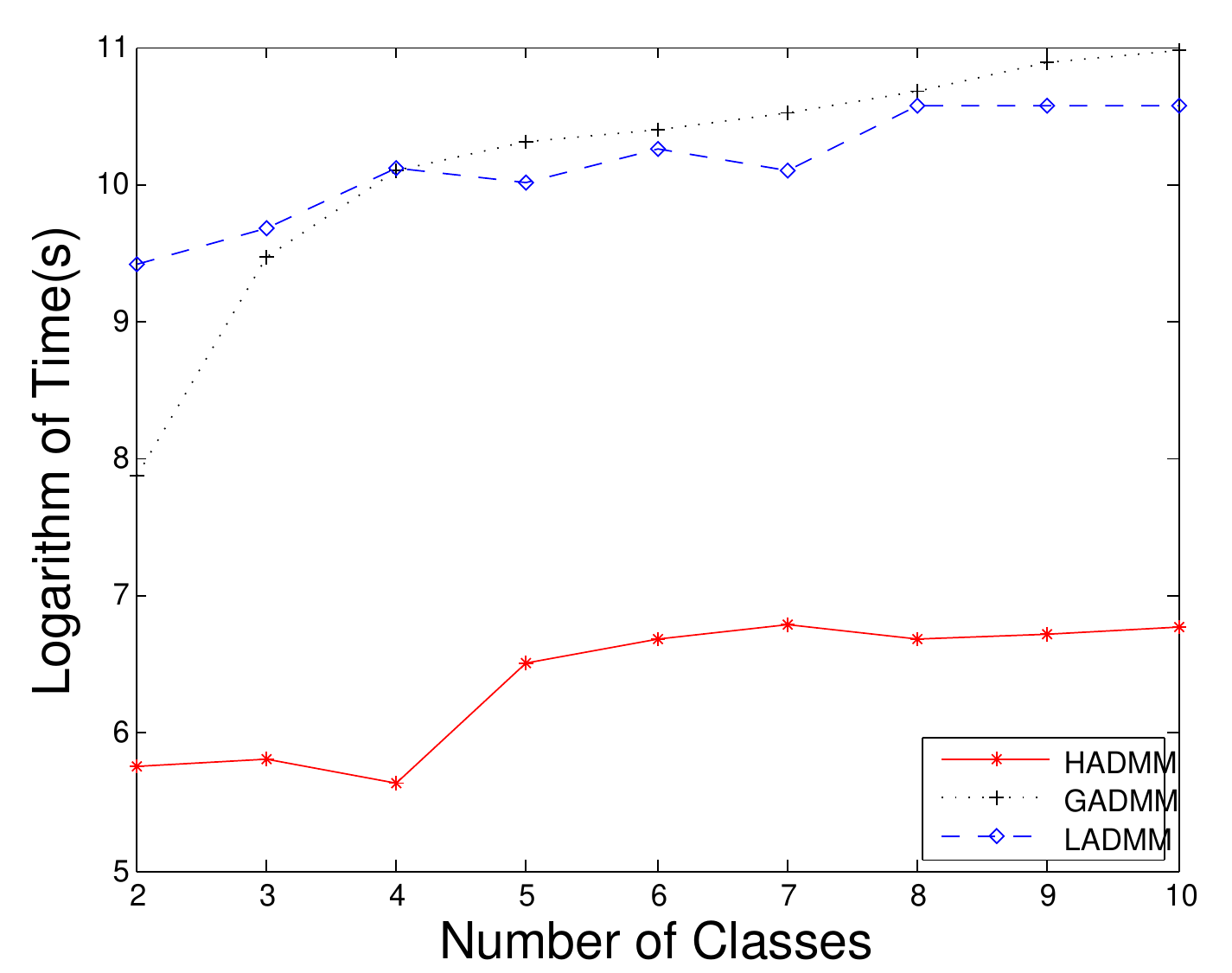}
 \includegraphics[width=0.31\textwidth]{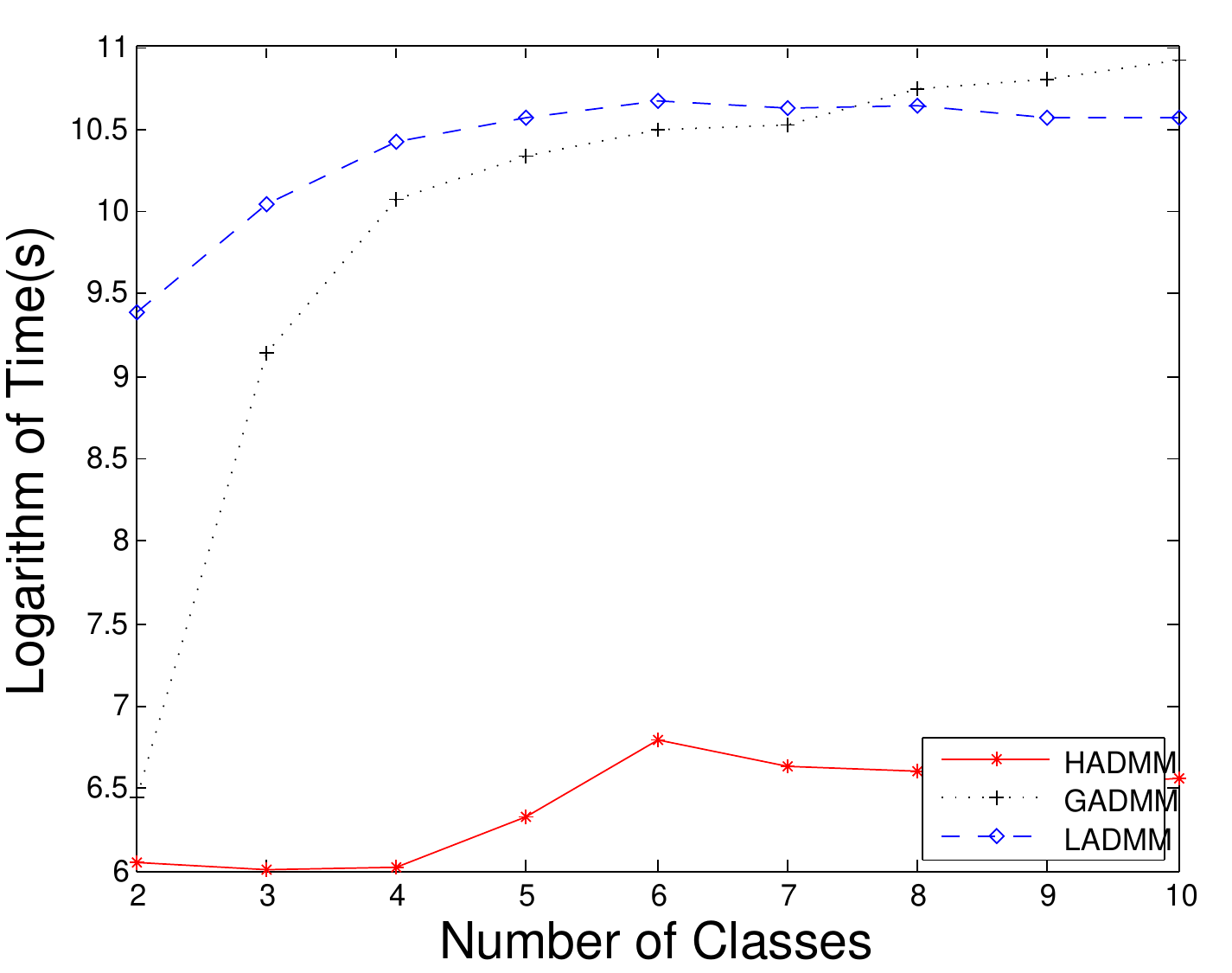}
 }
\caption{Logarithm of the running time (in seconds) of HADMM, LADMM and GADMM for $p = 1000$ on \textbf{Type A, Type B} and \textbf{Type C} data.}
\label{performance_simulation}
\end{figure*}

\begin{table}[h]
\caption{Objective function values of HADMM and ADMM on the six classes type C data (first 4 iterations, $p=1000$, $\lambda_1=0.0082$, $\lambda_2=0.0015$)}
\begin{center}
	\begin{tabular}{l | c | c | c | c }
	{\bf Iteration} & {\bf 1} &{\bf 2} &{\bf 3}	&{\bf 4} \\
	\hline
	ADMM 	& 1713.66 & -283.743 & -1191.94 & -1722.53 	  \\
	HADMM 	& 1734.42 & -265.073 & -1183.73 & -1719.78  	\\
	\end{tabular}
\end{center}
\label{precision_table}
\end{table}

To evaluate the correctness of our method HADMM, we compare the objective function value generated by HADMM to that by ADMM with respect to the number of iterations. We run the two methods for $500$ iterations over the three types of data with $p=1000$. As shown in Table \ref{precision_table}, in the first $4$ iterations, HADMM and ADMM yield slightly different objective function values. However, along with more iterations passed, both HADMM and ADMM converge to the same objective function value, as shown in Figure \ref{correctness} and Supplementary Figures S3-5. This experimental result confirms that our hybrid covariance thresholding algorithm is correct. We tested several pairs of hyper-parameters ($\lambda_1$ and $\lambda_2$) in our experiment. Please refer to the supplementary material for model selection.
Note that although in terms of the number of iterations HADMM and ADMM converge similarly, HADMM runs much faster than ADMM at each iteration, so HADMM converges in a much shorter time. 

\subsection{Performance on Synthetic Data}
In previous section we have shown that our HADMM converges to the same solution as ADMM. Here we test the running times of HADMM, LADMM and GADMM needed to reach the following stop criteria for $p=1000$: $\sum_{i=1}^{k}{||\bm{\Theta}^{(k)}-\bm{Y}^{(k)}||}<{10}^{-6}$ and $\sum_{i=1}^{k}{||\bm{Y}^{(k+1)}-\bm{Y}^{(k)}||}<{10}^{-6}$.
For $p=10000$, considering the large amount of running time needed for LADMM and GADMM, we run only 50 iterations for all the three methods and then compare the average running time for a single iteration. 

We tested the running time of the three methods using different parameters $\lambda_{1}$ and $\lambda_{2}$ over the three types of data. See supplementary material for model selection. We show the result for $p=1000$ in Figure \ref{performance_simulation} and that for $p=10000$ in Figure S15-23 in supplementary material, respectively. 

In Figure \ref{performance_simulation}, each row shows the experimental results on one type of data (\textbf{Type A}, \textbf{Type B} and \textbf{Type C} from top to bottom). Each column has the experimental results for the same hyper-parameters ($\lambda_1 = 0.009$ and $\lambda_2 = 0.0005$, $\lambda_1 = 0.0086$ and $\lambda_2 = 0.001$, and $\lambda_1 = 0.0082$ and $\lambda_2 = 0.0015$ from left to right). As shown in Figure \ref{performance_simulation}, HADMM is much more efficient than LADMM and GADMM. GADMM performs comparably to or better than LADMM when $\lambda_{2}$ is large. The running time of LADMM increases as $\lambda_{1}$ decreases. Also, the running time of all the three methods increases along with the number of classes. However, GADMM is more sensitive to the number of classes than our HADMM. Moreover, as our hybrid covariance thresholding algorithm yields finer non-uniform feasible partitions, the precision matrices are more likely to be split into many more smaller submatrices. This means it is potentially easier to parallelize HADMM to obtain even more speedup.

We also compare the three screening algorithms in terms of the estimated computational complexity for matrix eigen-decomposition, a time-consuming subroutine used by the ADMM algorithms. Given a partition $\mathcal{H}$ of the variable set of $\mathcal{V}$, the computational complexity can be estimated by $\sum_{\mathcal{H}_{i} \in \mathcal{H}} |\mathcal{H}_{i}|^{3}$. As shown in Supplementary Figures S6-14, when $p=1000$, our non-uniform thresholding algorithm generates partitions with much smaller computational complexity, usually $\frac{1}{10} \sim \frac{1}{1000}$ of the other two methods. Note that in these figures the Y-axis is the logarithm of the estimated computational complexity. When $p = 10000$, the advantage of our non-uniform thresholding algorithm over the other two are even larger, as shown in Figure S24-32 in Supplemental File.

\subsection{Performance on Real Gene Expression Data}

\begin{figure}[t]
\centering
\includegraphics[width=1.1\columnwidth]{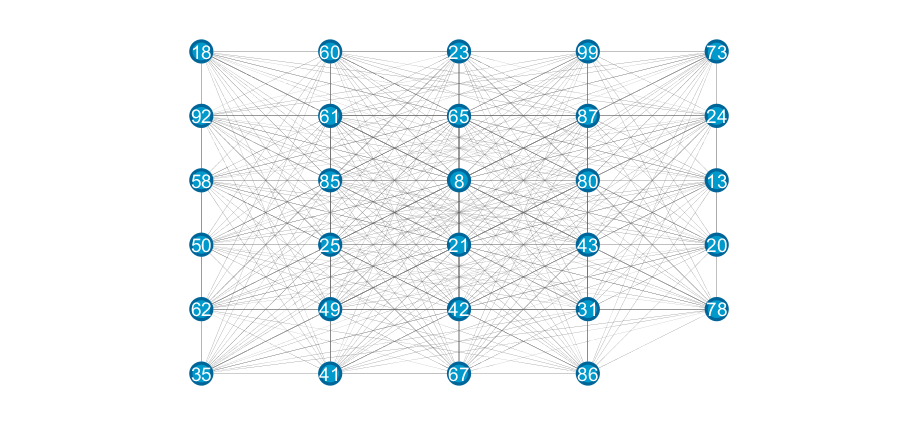}
\includegraphics[width=1.1\columnwidth]{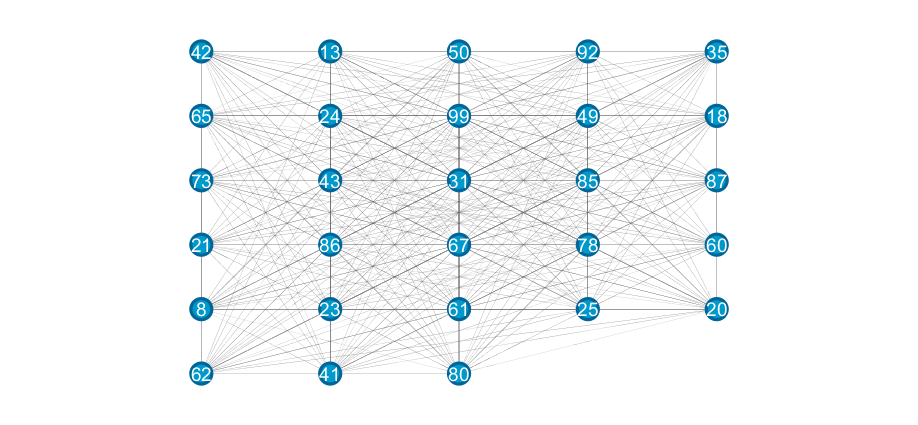}
\caption{Network of the first 100 genes of class one and class three for \textbf{Setting 1}.}
\label{gene_plot}
\end{figure}
We test our proposed method on real gene expression data. We use a lung cancer data (accession number GDS2771 \cite{spira2007airway}) downloaded from Gene Expression Omnibus and a mouse immune dataset described in \cite{jojic2013identification}. The immune dataset consists of 214 observations. The lung cancer data is collected from 97 patients with lung cancer and 90 controls without lung cancer, so this lung cancer dataset consists of two different classes: patient and control. We treat the $214$ observations from the immune dataset, the $97$ lung cancer observations and the $90$ controls as three classes of a compound dataset for our joint inference task. These three classes share $10726$ common genes, so this dataset has $10726$ features and $3$ classes. As the absolute value of entries of covariance matrix of first class (corresponds to immune observations) are relatively larger, so we divide each entry of this covariance matrix by $2$ to make the three covariance matrices with similar magnitude before performing joint analysis using unique $\lambda_1$ and $\lambda_2$. \par 
The running time (first $10$ iterations) of HADMM, LADMM and GADMM for this compound dataset under different settings are shown in Table \ref{compare_over_gene} and the resultant gene networks with different sparsity are shown in Fig \ref{gene_plot} and Supplemental File. \par 
As shown in Table \ref{compare_over_gene}, HADMM (ADMM + our hybrid screening algorithm) is always more efficient than the other two methods in different settings. Typically, when $\lambda_1$ is small and $\lambda_2$ is large (\textbf{Setting 1}), our method is much faster than LADMM. In contrast, when $\lambda_2$ is small and $\lambda_1$ is large enough (\textbf{Setting 4} and \textbf{Setting 5}), our method is much faster than GADMM. What's more, when both $\lambda_1$ and $\lambda_2$ are with moderate values (\textbf{Setting 2} and \textbf{Setting 3}), HADMM is still much faster than both GADMM and LADMM. 
\begin{table}[h!]
\caption{Running time (hours) of HADMM, LADMM and GADMM on real data. (\textbf{Setting 1}: $\lambda_{1} = 0.1$ and $\lambda_{2} = 0.5$; \textbf{Setting 2}: $\lambda_{1} = 0.2$ and $\lambda_{2} = 0.2$; \textbf{Setting 3}: $\lambda_{1} = 0.3$ and $\lambda_{2} = 0.1$; \textbf{Setting 4}: $\lambda_{1} = 0.4$ and $\lambda_{2} = 0.05$, and \textbf{Setting 5}: $\lambda_{1} = 0.5$ and $\lambda_{2} = 0.01$)}
\begin{center}
	\begin{tabular}{l | c | c | c | c | c}
	{\bf Method} & {\bf Setting 1} &{\bf Setting 2} &{\bf Setting 3} &{\bf Setting 4} &{\bf Setting 5}	\\
	\hline
	HADMM 	& 3.46 & 8.23 & 3.9 & 1.71 & 1.11	  \\
	LADMM 	& $>$ 20 & $>$ 20 & 13.6 & 3.72 & 1.98 	\\
	GADMM 	& 4.2 & $>$ 20 & $>$ 20 & 11.04 & 6.93  \\
	\end{tabular}
\end{center}
\label{compare_over_gene}
\end{table}

As shown in Fig \ref{gene_plot}, the two resultant networks are with very similar topology structure. This is reasonable because we use large $\lambda_2$ in \textbf{Setting 1}. Actually, the networks of all the three classes under \textbf{Setting 1} share very similar topology structure. What's more, the number of edges in the network does decrease significantly as $\lambda_1$ goes to $0.5$, as shown in Supplementary material.

\section{Conclusion and Discussion}
This paper has presented a non-uniform or hybrid covariance thresholding algorithm to speed up solving group graphical lasso. We have established necessary and sufficient conditions for this thresholding algorithm. Theoretical analysis and experimental tests demonstrate the effectiveness of our algorithm. Although this paper focuses only on group graphical lasso, the proposed ideas and techniques may also be extended to fused graphical lasso.

In the paper, we simply show how to combine our covariance thresholding algorithm with ADMM to solve group graphical lasso. In fact, our thresholding algorithm can be combined with other methods developed for (joint) graphical lasso such as the QUIC algorithm \cite{hsieh2011sparse}, the proximal gradient method \cite{rolfs2012iterative}, and even the quadratic method developed for fused graphical lasso \cite{yang2012fused}.

The thresholding algorithm presented in this paper is static in the sense that it is applied as a pre-processing step before ADMM is applied to solve group graphical lasso. We can extend this ``static'' thresholding algorithm to a ``dynamic'' version. For example, we can identify zero entries in the precision matrix of a specific class based upon intermediate estimation of the precision matrices of the other classes. By doing so, we shall be able to obtain finer feasible partitions and further improve the computational efficiency.

\nocite{langley00}
\bibliography{SCREEN}
\bibliographystyle{splncs03}

\end{document}